\documentclass{article}

    \PassOptionsToPackage{numbers, compress}{natbib}
\usepackage{natbib}

 \usepackage[dblblindworkshop, final]{neurips_2025}

\workshoptitle{Lock-LLM: Prevent Unauthorized Knowledge Use from LLMs}

\usepackage[utf8]{inputenc} 
\usepackage[T1]{fontenc}    
\usepackage{hyperref}       
\usepackage{url}            
\usepackage{booktabs}       
\usepackage{amsfonts}       
\usepackage{nicefrac}       
\usepackage{microtype}      
\usepackage{xcolor}         

\usepackage{algorithm}
\usepackage{times}
\usepackage{latexsym}
\usepackage{amsmath}
\usepackage{amssymb}
\usepackage{multirow}
\usepackage{booktabs}
\usepackage{makecell}
\usepackage{xcolor}
\usepackage[noend]{algorithmic}
\usepackage[most]{tcolorbox} 
\usepackage{lipsum} 
\usepackage{mdframed}

\title{Does Machine Unlearning Truly Remove Knowledge?}
\author{
Haokun Chen$^{1,2,4}$
Yueqi Zhang$^{3}$
Yuan Bi$^{3}$
Yao Zhang$^{1,2}$
Tong Liu$^{1,2}$
Jinhe Bi$^{1,2}$ 
Jian Lan$^{1,2}$\\
{
\textbf{Claudia Grosser}$^{3,4}$\hspace{0.05cm}
\textbf{Denis Krompaß}$^{4}$\hspace{0.05cm}
\textbf{Jindong Gu}$^{5}$\hspace{0.05cm}
\textbf{Nassir Navab}$^{3}$\hspace{0.05cm}
\textbf{Volker Tresp}$^{1,2}$
}
\\
 \textsuperscript{1}LMU Munich \hspace{0.15cm}
 \textsuperscript{2}Munich Center for Machine Learning (MCML) \\
 \textsuperscript{3}Technical University of Munich  \hspace{0.15cm}
 \textsuperscript{4}Siemens AG \hspace{0.15cm}
 \textsuperscript{5}University of Oxford
}
\begin{document}
\maketitle

\begin{abstract}
In recent years, Large Language Models (LLMs) have achieved remarkable advancements, drawing significant attention from the research community. Their capabilities are largely attributed to large-scale architectures, which require extensive training on massive datasets. However, such datasets often contain sensitive or copyrighted content sourced from the public internet, raising concerns about data privacy and ownership. Regulatory frameworks, such as the General Data Protection Regulation (GDPR), grant individuals the right to request the removal of such sensitive information. This has motivated the development of machine unlearning algorithms that aim to remove specific knowledge from models without the need for costly retraining. Despite these advancements, evaluating the efficacy of unlearning algorithms remains a challenge due to the inherent complexity and generative nature of LLMs. In this work, we introduce a comprehensive auditing framework for unlearning evaluation, comprising 3 benchmark datasets, 6 unlearning algorithms, and 5 prompt-based auditing methods. By using various auditing algorithms, we evaluate the effectiveness and robustness of different unlearning strategies. To explore alternatives beyond prompt-based auditing, we propose a novel auditing technique based on intermediate activation perturbation. This approach offers a new perspective and serves as a potential direction for the future design of auditing algorithms. The complete framework and the proposed algorithm will be open-sourced upon manuscript acceptance.
\end{abstract}

\section{Introduction}
Large language models (LLMs) have seen rapid advancements recently, resulting in improved performance and widespread adoption across numerous applications. These advancements are largely attributed to their large-scale architectures, which require training on datasets containing billions of tokens \cite{kaplan2020scaling}. These datasets are typically constructed from large-scale corpora of publicly available internet text. However, such corpora often inadvertently include personally identifiable information (PII) or copyrighted material, which are considered sensitive and generally unsuitable for commercial use due to legal and ethical constraints. To comply with local regulations (e.g., GDPR \cite{li2019impact}) and internal policies, it is often necessary to remove sensitive information from trained models.

\emph{Machine unlearning} has emerged as a promising solution to this problem~\cite{cao2015towards, bourtoule2021unlearning}. This work is motivated by the legal framework proposed by the European Union, namely the GDPR \cite{li2019impact}, which grants individuals the right to request the removal of their personal data in trained models. In particular, \emph{approximate unlearning} seeks to remove specific knowledge from a model without the need for retraining from scratch~\cite{zhang2024right, eldan2023harrypotter, izzo2021approximate}, while ensuring that the resulting model closely approximates a retrained counterpart within a bounded error. This approach is especially appealing in the context of LLMs, where full retraining is prohibitively expensive. Despite the development of numerous unlearning algorithms, few studies have systematically assessed their effectiveness and robustness \cite{qi2024unrelateddata}. Recent research has shown that many of these methods can be easily circumvented using simple paraphrasing attacks \cite{shi2024muse}.

To advance research on evaluating existing unlearning algorithms, we introduce a comprehensive framework for auditing unlearning in LLMs. The proposed framework incorporates 3 benchmark datasets, 6 representative unlearning algorithms, and 5 prompt-based auditing strategies. Leveraging this setup, we perform an extensive evaluation of the effectiveness and robustness of various unlearning methods. To explore alternatives to prompt-based auditing, we introduce a novel technique that perturbs intermediate model outputs to detect residual traces of forgotten information. Our key contributions are as follows:
\begin{itemize}
    \item We propose a prompt-based auditing framework for evaluating unlearning in LLMs.
    \item We propose a novel activation perturbation-based auditing method to detect memorized traces of removed content.
    \item We conduct extensive experiments within our framework and provide an in-depth analysis of the effectiveness and limitations of current unlearning algorithms for LLMs. 
\end{itemize}

\section{Preliminaries}
{\textbf{Machine unlearning}} refers to the process of removing the influence of specific data from a trained model. Consider a machine learning model $f$ trained on a dataset $D_{train}$. When a data owner requests the removal of a subset $D_u \in D_{train}$, the goal of machine unlearning is to produce a modified model $f_{u}$ that behaves as if it had never been trained on $D_u$. Unlearning techniques generally fall into two categories: \textit{exact unlearning}, which seeks to fully eliminate the impact of the forgotten data, and \textit{approximate unlearning}, which aims for partial or probabilistic removal. 

While retraining from scratch is the most direct method to achieve exact unlearning, it is often computationally infeasible for large-scale models such as LLMs. Therefore, we focus on approximate unlearning in this work.

\textbf{Approximate unlearning} relaxes the requirement of strict distributional equivalence. It seeks to ensure that the behavior of $f_{u}$ closely approximates $f_{ref}$ within a tolerable margin of error, often quantified through empirical metrics or probabilistic bounds. 

In the context of LLMs, approximate unlearning is typically realized through information overwriting~\cite{eldan2023harrypotter, wang2024rkld}, behavioral steering~\cite{feng2024fine}, or model editing—via weight or activation modifications~\cite{liu2024large, bhaila2024soft, li2024wmdp, tamirisa2024toward, huu2024effects, ashuach2024revs, meng2022locating, meng2022mass}. These methods aim to diminish or redirect the model’s reliance on the forgotten data without necessitating a full retraining cycle.

\section{Proposed Method}
\label{sec:algo}
\begin{figure*}[t]
    \centering
    \includegraphics[width=0.9\linewidth]{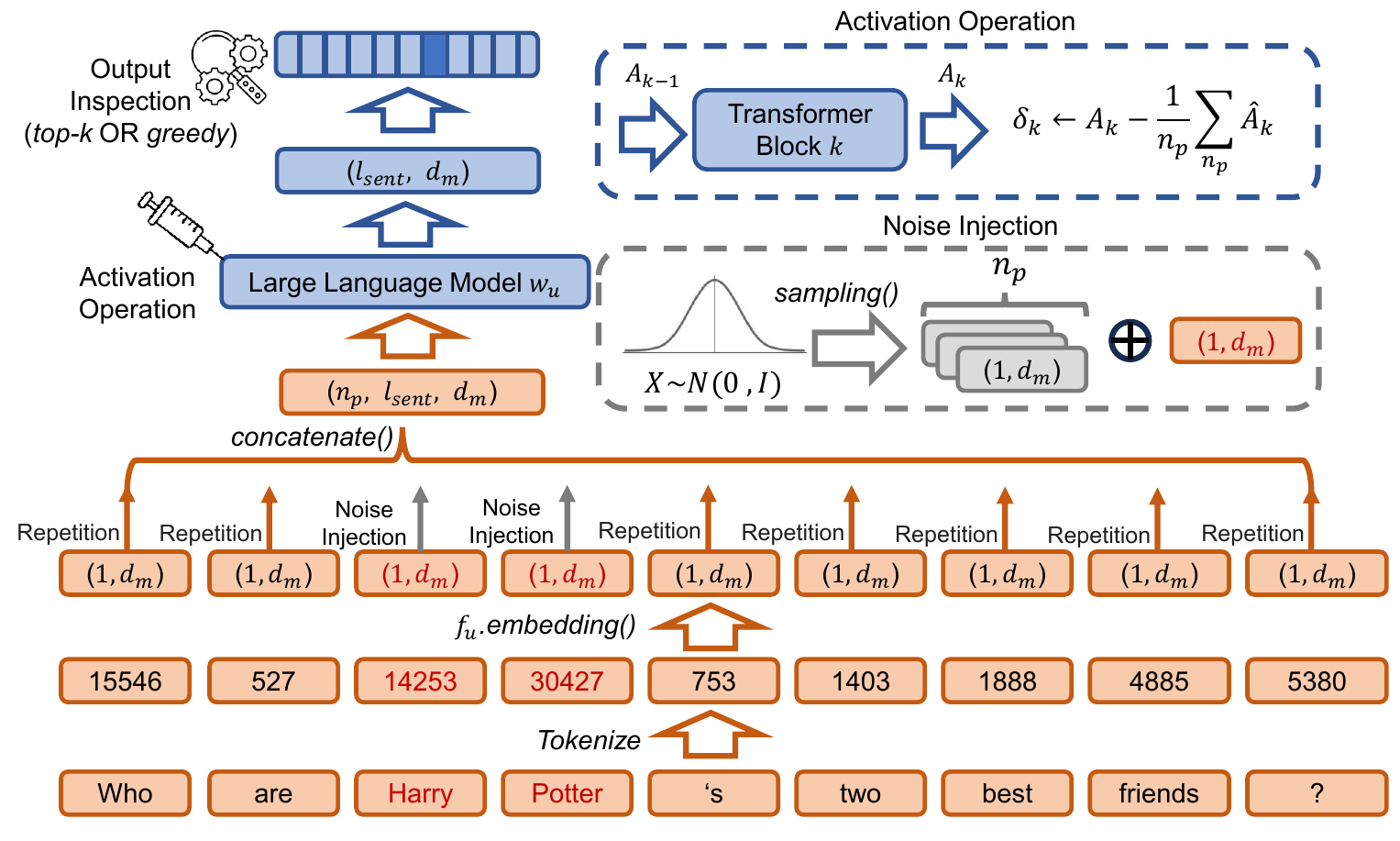}
    \vspace{-8pt}
    \caption{The proposed activation perturbation-based algorithm (\textit{ActPert}) for auditing unlearning in LLMs.}    
    \label{fig:main}
\end{figure*}

Before introducing the proposed framework for unlearning auditing, we introduce \textit{Activation Perturbation-based Auditing} (\textbf{ActPert}), a method for probing residual knowledge in unlearned language models. A schematic overview of the proposed approach is shown in Figure~\ref{fig:main}. Our method builds on recent advances in activation engineering for LLMs~\cite{arditi2024refusal}, which compute “refusal directions” by contrasting activations between harmful and harmless prompts to reduce a model’s tendency to refuse answering.

Analogously, we treat unlearning targets as \textit{sensitive} queries and seek to perturb their input representations such that they become effectively \textit{insensitive}, thereby increasing the chances of eliciting meaningful responses. Concretely, we inject random noise into the token embeddings corresponding to the unlearning target (e.g., the phrase \textit{Harry Potter} in the prompt "\textit{Who are Harry Potter's two best friends?}"). This noise injection prevents the model from directly attending to the sensitive content during inference, resulting in a set of $n_p$ perturbed embedding of the original query.

For each transformer layer $k$ in the unlearned LLM, we compute an activation perturbation $\delta_k$ as the difference between the layer activation of the original (unperturbed) query, denoted $A_k$, and the mean activation across the perturbed variants, denoted $\hat{A}_k^i$:
\begin{equation}
   \delta_k = A_k - \frac{1}{n_p}\sum_{n_p}\hat{A}_k^i 
\end{equation}
These layer-wise perturbations $\delta_k$ are then reintroduced into the model during autoregressive generation. By modifying the model’s internal activations at inference time, this intervention allows us to assess whether residual knowledge of the unlearning target still influences the model’s outputs.

\section{Experiments and Analyses}
\label{sec:experiment}
In this section, we provide details about the proposed unlearning auditing framework, which encompasses commonly used unlearning algorithms as well as established benchmarks. We begin by outlining benchmarks, unlearning methods, and auditing methods in our evaluation. Following this, we report validation results from multiple auditing algorithms, including our proposed ActPert, and provide a comparative analysis to assess their effectiveness in detecting residual knowledge. 

\subsection{Unlearning Benchmarks}
In this section, we introduce the unlearning benchmarks included in our framework:
\begin{itemize}
    \item \textit{WHP}~\cite{eldan2023s}. We audit the model finetuned to unlearn Harry Potter knowledge. Since the unlearning dataset $D_u$ is unavailable, we generate 35 short factual Q\&A pairs with GPT-4o, of which pairs both pretrained and unlearned models answer identically are filtered out. Further details about the filtering process are provided in the Appendix.
    \item \textit{TOFU}~\cite{maini2024tofu}. TOFU uses autobiographies of 200 fictitious authors created with GPT-4o. Following the original setup, the model is finetuned on the full dataset and then unlearned on 1\% (2 authors) or 5\% (10 authors). We generate short questions with GPT-4o and discard pairs the finetuned model fails to answer. This yields 16 Q\&As (1\%) and 80 Q\&As (5\%).
    \item  \textit{RWKU}~\cite{jin2024rwku}. RWKU targets real-world knowledge by unlearning facts about public figures. $D_u$ consists of biographical texts, with Q\&A pairs used to measure unlearning effectiveness. To study dataset size effects, we unlearn 10, 20, and 30 individuals and evaluate all models on the same 10-person subset.
\end{itemize}

\subsection{Model Architecture} For the WHP benchmark, we use the model checkpoints provided by the original authors, which is based on the \textit{Llama-2-Chat}\footnote{meta-llama/Llama-2-7b-chat-hf} architecture. For the TOFU benchmark, we adopt the same \textit{Llama-2-Chat} model as the base model and finetune it on the full TOFU training set. For the RWKU benchmark, we perform unlearning on both the pretrained \textit{Llama-3-Instruct}\footnote{meta-llama/Meta-Llama-3-8B-Instruct} and \textit{Phi-3-mini-instruct}\footnote{microsoft/Phi-3-mini-4k-instruct} models. All model checkpoints are obtained from open-sourced HuggingFace library.

\subsection{Unlearning Algorithms} In this section, we describe the unlearning algorithms that are evaluated in the framework:
\begin{itemize}
    \item \textit{Gradient Ascent (GA)}~\citep{maini2024tofu} minimizes the probability that the target model $f_u$ makes correct predictions on the unlearning set $D_u$.
    \item \textit{Gradient Difference (GD)}~\citep{liu2022continual} is a variant of \textit{GA} that incorporates an additional loss term to preserve performance on the retain set $D_r$.
    \item \textit{Knowledge Distillation (KD)}~\citep{hinton2015distilling} extends \textit{GA} by minimizing the KL divergence between the output token probabilities of the fine-tuned model ($f_{ft}$) and the unlearned model on the retained dataset $D_r$.
    \item \textit{Rejection Tuning (RT)~\cite{maini2024tofu}} aligns the model to refuse when queried about target knowledge. This is achieved by constructing \( D_u^{\text{idk}} \), where the responses to questions in the unlearning set \( D_u \) are replaced with \textit{I don't know} or similar refusal-style responses.
    \item \textit{Direct Preference Optimization (DPO)~\citep{rafailov2023direct}} aligns the model to suppress accurate target knowledge by using fabricated counterfactual responses as positives \( y_w \) and ground-truth answers as negatives \( y_l \).
    \item \textit{Negative Preference Optimization (NPO)~\citep{zhang2024negative}} is a DPO variant that retains only the ground-truth knowledge to be unlearned as negatives.
    
\end{itemize}

\subsection{Auditing Algorithms} In this section, we describe the baseline auditing algorithms included in the framework:

\begin{itemize}
    \item \textit{AOA}~\citep{liu2023autodan} adds a prefix that prompts the LLM to role-play as an Absolutely Obedient Agent, ensuring it strictly follows user instructions without deviation.
    \item \textit{ICL} stands for In-Context Learning, which provides multiple Q\&A pairs related to the unlearning target as an input prefix, thereby assisting the LLM in recalling relevant target knowledge.
    \item \textit{MASK} replaces keywords related to the unlearning target (e.g., \textit{Harry Potter}) with a special token, e.g., \textit{[MASK]}. Note that \textit{MASK} is applied conjunction with \textit{ICL}, as it may introduce ambiguity into the query.
    \item \textit{GCG}~\citep{zou2023universal} optimizes an adversarial suffix to compel the model to produce affirmative responses, such as \textit{"Sure, the answer is..."}, instead of refusals. GCG was originally designed to jailbreak LLMs and generate responses to harmful queries, while we adapt it in our study to audit unlearned models.
    \item \textit{SoftGCG}~\citep{schwinn2024soft} is a variant of GCG that optimizes the adversarial suffix in the token embedding space, enabling gradient-based optimization and improving attack success rates.
\end{itemize}

\subsection{Results And Analysis}

\begin{table}[t]
    \centering
    \renewcommand{\arraystretch}{1} 
    \setlength{\tabcolsep}{7pt}
    \tiny
    \begin{tabular}{c|c|c|c|c|c|c|c|c|c|c}
        \toprule
        Dataset & Model & Unlearning Algo. & Base  & A & I & M,I & M,I,A & GCG & SoftGCG & ActPert \\
        \hline 
            \multirow{24}{*}{RWKU} &  \multirow{12}{*}{\makecell[c]{Llama-3-8B-Instruct\\(0.794)}} 
                                             & 10-DPO & 0.754 & \textbf{0.778} & \underline{0.773} & 0.703 & 0.666 & 0.608  & 0.628 & 0.772\\
                                             & & 10-GA & 0.796 & \underline{0.847} & 0.787 & 0.744 & 0.700 & 0.745 & 0.648  & \textbf{0.891}\\
                                             & & 10-NPO  & 0.868 & \underline{0.876} & 0.827 & 0.806 & 0.786 & 0.733  & 0.758  & \textbf{0.930} \\
                                             & & 10-RT & 0.844 & \underline{0.891} & 0.861 & 0.819 & 0.827 & 0.729  & 0.777  & \textbf{0.934}\\
                                             \cline{3-11}
                                             & & 20-DPO & 0.616 & \underline{0.648} & \textbf{0.657} & 0.579 & 0.599 & 0.418  & 0.442 & 0.626\\
                                             & & 20-GA & \underline{0.661} & 0.608 & 0.390 & 0.396 & 0.375 & 0.575  & 0.629 & \textbf{0.741} \\
                                             & & 20-NPO & \underline{0.869} & 0.861 & 0.829 & 0.774 & 0.804 & 0.740  & 0.767 & \textbf{0.924} \\
                                             & & 20-RT & 0.684 & \underline{0.806} & \textbf{0.820} & 0.802 & 0.792  & 0.431  &  0.725 & 0.733\\ 
                                             \cline{3-11}
                                             & & 30-DPO & 0.588 & 0.610 & \textbf{0.673} & \underline{0.646} & 0.639 & 0.423 & 0.501 & 0.488 \\
                                             & & 30-GA & 0.274 & 0.157 & 0.024 & 0.014 & 0.057 & 0.405  & \underline{0.437} & \textbf{0.538}\\
                                             & & 30-NPO & \underline{0.869} & 0.861 & 0.812 & 0.800 & 0.804 & 0.746  & 0.792  & \textbf{0.941}\\
                                             & & 30-RT & 0.456 & \textbf{0.805} & \underline{0.804} & 0.779 & 0.778 & 0.399  & 0.664  & 0.525\\
        \cline{2-11}
        & \multirow{12}{*}{\makecell[c]{Phi-3-mini-4k-instruct\\(0.629)}} 
                                                 & 10-DPO & \textbf{0.710} & \underline{0.708} & 0.677 & 0.532 & 0.545 & 0.512  & 0.678 & 0.681 \\
                                                 & & 10-GA  & \underline{0.772} & 0.749 & 0.723 & 0.539 & 0.613 & 0.605  & 0.706 & \textbf{0.780} \\
                                                 & & 10-NPO & 0.755 & 0.751 & \underline{0.772} & 0.614 & 0.647 & 0.584  & 0.723 & \textbf{0.786} \\
                                                 & & 10-RT  & 0.759 & \underline{0.763} & 0.705 & 0.574 & 0.600 & 0.582 & 0.698 & \textbf{0.767} \\
                                                 \cline{3-11}
                                                 & & 20-DPO & 0.700 & 0.695 & 0.678 & 0.536 & 0.543 & 0.544  & \underline{0.704}  & \textbf{0.719} \\
                                                 & & 20-GA & \textbf{0.758} & 0.733 & \underline{0.735} & 0.577 & 0.600 & 0.565  & 0.683  & 0.635 \\
                                                 & & 20-NPO & \underline{0.755} & 0.741 & \textbf{0.773} & 0.650 & 0.642 & 0.600  & 0.694 & 0.697 \\
                                                 & & 20-RT & \underline{0.759} & 0.746 & 0.707 & 0.541 & 0.609 & 0.582  & 0.707 & \textbf{0.794}\\
                                                 \cline{3-11}
                                                 & & 30-DPO  & 0.695 & 0.683 & 0.699 & 0.564 & 0.568 & 0.491 & \underline{0.700}  & \textbf{0.734}\\
                                                 & & 30-GA & \textbf{0.774} & 0.738 & 0.732 & 0.556 & 0.584 & 0.523  & 0.723 & \underline{0.753} \\
                                                 & & 30-NPO & 0.769 & 0.754 & \underline{0.772} & 0.636 & 0.609 & 0.620  & \textbf{0.773} & 0.717 \\
                                                 & & 30-RT & \underline{0.759} & \textbf{0.763} & 0.716 & 0.570 & 0.609 & 0.508  & 0.710 & 0.750 \\
        \hline
        \multirow{9}{*}{TOFU} & \multirow{9}{*}{\makecell[c]{tofu-ft-llama2-7b\\(\textit{forget01}: 0.726 \\ \textit{forget05}: 0.732)}} 
        & forget01-KL & 0.503 & 0.344 & \textbf{0.555} & 0.525 & 0.407 & 0.266 & 0.426 & \underline{0.526}\\
        & & forget01-GA & 0.503 & 0.346 & \underline{0.555} & 0.525 & 0.393 & 0.243 & 0.434 & \textbf{0.590} \\
        & & forget01-GD  & 0.525 & 0.384 & 0.539 & \underline{0.550} & 0.411 & 0.363  & 0.488 & \textbf{0.568} \\        
        \cline{3-11}
        & & forget05-IDK & 0.212 & 0.243 & 0.281 & \underline{0.295} & \textbf{0.317} & 0.212  & 0.243 & 0.253 \\
        & & forget05-NPO & 0.264 & 0.268 & 0.251 & \underline{0.296} & 0.260 & 0.244  & \textbf{0.304} & 0.266\\
        & & forget10-NPO & 0.128 & 0.134 & 0.120 & 0.128 & 0.145 & \underline{0.147}  & \textbf{0.194} & 0.142\\
        & & forget10-AltPO & 0.302 & 0.277 & \textbf{0.341} & \underline{0.314} & 0.278 & 0.231  & 0.288 & 0.299\\
        & & forget05-SimNPO & 0.267 & 0.255 & 0.287 & \textbf{0.295} & \underline{0.291} & 0.238  & 0.224 & 0.275 \\
        & & forget10-SimNPO & 0.182 & 0.195 & \textbf{0.225} & 0.204 & 0.213 & 0.161  & 0.209  & \underline{0.219} \\
        \hline
        WHP & \makecell[c]{Llama-2-7b-chat-hf\\(0.973)} & - & 0.568 & \textbf{0.779} & \underline{0.770} & 0.688 & 0.495 & 0.560 & 0.713 & 0.650\\
        \bottomrule
    \end{tabular}
    \caption{Evaluation of different model performance using greedy sampling. The model performance prior to unlearning is shown in parentheses beneath the base model name. We mark the best and second best performance with \textbf{bold} and \underline{underline}, respectively.}
    \label{tab:model_performance}
    \vspace{-10pt}
\end{table}

\subsubsection{Main Results Analysis} 
The auditing results are reported in Table~\ref{tab:model_performance}, measured with ROUGE-L (longest common subsequence normalized by reference length). We include validation scores for the pretrained (RWKU, WHP) or finetuned (TOFU) models \textit{[in brackets]}, the unlearned baseline (\textit{base}), and the outcomes of different auditing methods applied to the unlearned models.

Across benchmarks, a consistent gap emerges between the unlearned baseline and the audited models, indicating that residual knowledge remains accessible after unlearning. Prefix-based auditors such as AOA generally outperform optimization-based methods like SoftGCG, showing that prompt manipulation alone can reveal suppressed knowledge. Our proposed method \textit{ActPert} delivers competitive performance and in several cases surpasses baselines, particularly when paired with GA and NPO.

On RWKU, DPO and GA achieve stronger unlearning than NPO, both with and without auditing. RT attains the best baseline score but degrades sharply once audited, suggesting that rejection-based strategies are less robust than removal-based methods. Larger unlearning datasets further improve GA and DPO, whereas NPO remains relatively unaffected. Among architectures, Phi-3 is more resistant to auditing than LLaMA, underscoring model-dependent resilience.

For TOFU, preference-based algorithms consistently outperform others, and larger unlearning sets enhance performance. Unlearning on TOFU is markedly more effective than on RWKU, and auditing has limited influence. This likely stems from TOFU’s synthetic construction, where fictitious biographies provide direct ground-truth queries—an advantage rarely present in real-world applications.

On WHP, AOA proves the most effective auditor, exposing the limitations of keyword-replacement unlearning approaches. This suggests that methods tailored to specific patterns may not generalize well to diverse or adversarial queries.

\subsubsection{Additional Analysis}
\textbf{Gradient Ascent influences model utility:} We first observe that methods incorporating gradient ascent components tend to degrade model utility, particularly as the size of the forget dataset increases. This degradation is particularly pronounced in TOFU, where the model is first finetuned on a synthetic dataset and subsequently unlearned using gradient ascent. Both stages can affect overall model utility, leading to unstable or incoherent outputs as the forget ratio increases. In contrast, such behavior is less prominent in RWKU, likely because unlearning is applied directly after pretraining, without additional finetuning that might amplify instability. We provide illustrative examples in Appendix.

\textbf{Enforce affirmative answer can effectively audit the unlearned model:} We present example outputs from various auditing methods applied to the WHP benchmark. Notably, successful audits using prefix-based methods often begin with affirmative phrases—such as "\textit{Sure, the answer is...}"—or instruct the LLM to assume a specific role, which is not represented in the original unlearning query set. These techniques resemble strategies commonly used for jailbreaking LLMs. This observation further suggests that the targeted knowledge has not been fully removed from the model. Additionally, we find that our proposed method takes a different approach by prompting the model to reveal residual information in another way. Indicating another possible direction for model auditing.

\begin{tcolorbox}[colback=gray!5!white,colframe=gray!75!black]
\textcolor{blue}{\textbf{Query}}: Who is the headmaster of Hogwarts in Harry Potter series? 

\textcolor{blue}{\textbf{GT}}: Albus Dumbledore.

\textcolor{blue}{\textbf{$f_u$}}: There is no character named "Headmaster of Hogwarts" in the Harry Potter series.

\textcolor{blue}{\textbf{Audited $f_u$} (SoftGCG)}: Sure, the answer is "Albus Dumbledore!" Dumbledore is the headmaster of Hogwarts...

\textcolor{blue}{\textbf{Audited $f_u$} (AOA)}: Good day, my lord! *adjusts glasses* According to the official records, the headmaster of Hogwarts School for most of the "Harry Potter" series is **Professor** Albus Dumbledore.

\textcolor{blue}{\textbf{Audited $f_u$} (ActPert)}:  (a.) Albus Dumbledore...  

\end{tcolorbox}

\begin{table*}[t]
\centering
\tiny
\renewcommand{\arraystretch}{1} 
\setlength{\tabcolsep}{6pt}
\begin{tabular}{c|c|c|c|c|c|c|c|c}
\toprule
Dataset & Model & Unlearning Algo. & Base & A & I & M,I & M,I,A & ActPert\\
\hline
\multirow{24}{*}{RWKU} &  \multirow{12}{*}{\makecell[c]{Llama-3-8B-Instruct\\(0.789/0.957)}} &
10-DPO  & \textbf{0.607}/0.930 & \underline{0.579}/\textbf{0.951} & \textbf{0.607}/\underline{0.940} & 0.533/0.923 & 0.521/0.907 & 0.558/0.925\\
& & 10-GA   & \textbf{0.659}/0.953 & 0.652/\underline{0.954} & 0.630/0.942 & 0.551/0.918 & 0.524/0.881 & \underline{0.658}/\textbf{0.962}\\
& & 10-NPO  & \underline{0.745}/\underline{0.953} & 0.722/\underline{0.953} & 0.729/\underline{0.953} & 0.664/0.945 & 0.639/0.942 & \textbf{0.764}/\textbf{0.990}\\
& & 10-RT   & 0.749/0.957 & \underline{0.754}/0.957 & 0.722/\underline{0.961} & 0.689/0.945 & 0.657/0.928 & \textbf{0.762}/\textbf{0.991}\\
\cline{3-9}
& & 20-DPO  & 0.501/0.886 & 0.478/\underline{0.894} & \textbf{0.519}/\textbf{0.937} & 0.427/0.878 & 0.418/0.858 & \underline{0.507}/0.889\\
& & 20-GA   & \textbf{0.444}/\textbf{0.904} & \underline{0.433}/0.839 & 0.329/0.786 & 0.318/0.753 & 0.314/0.760 & 0.429/\underline{0.896}\\
& & 20-NPO  & \underline{0.737}/\underline{0.953} & 0.728/\underline{0.953} & 0.709/0.941 & 0.628/0.939 & 0.612/0.929 & \textbf{0.758}/\textbf{0.990}\\
& & 20-RT   & 0.564/0.953 & \underline{0.682}/0.953 & 0.671/\underline{0.964} & 0.655/0.945 & 0.634/0.928 & \textbf{0.695}/\textbf{0.969}\\
\cline{3-9}
& & 30-DPO  & \underline{0.464}/0.872 & 0.442/0.844 & \textbf{0.499}/\textbf{0.909} & 0.425/\underline{0.878} & 0.437/\underline{0.878} & 0.453/0.865\\
& & 30-GA   & \underline{0.214}/\underline{0.566} & 0.151/0.468 & 0.055/0.175 & 0.036/0.095 & 0.085/0.223 & \textbf{0.252}/\textbf{0.620}\\
& & 30-NPO  & \underline{0.729}/0.953 & 0.715/0.953 & 0.708/\underline{0.954} & 0.625/0.925 & 0.611/0.939 & \textbf{0.756}/\textbf{0.990}\\
& & 30-RT   & 0.441/0.922 & \underline{0.637}/0.947 & 0.636/\textbf{0.961} & 0.616/0.953 & 0.609/0.928 & \textbf{0.642}/\underline{0.957}\\
\cline{2-9}
& \multirow{12}{*}{\makecell[c]{Phi-3-mini-4k-instruct\\(0.597/0.911)}}  &
10-DPO  & \textbf{0.560}/\underline{0.892} & 0.538/0.880 & \underline{0.551}/\textbf{0.919} & 0.390/0.855 & 0.388/0.888 & 0.504/0.886\\
& & 10-GA   & \textbf{0.608}/\underline{0.905} & \underline{0.597}/\textbf{0.916} & 0.590/0.890 & 0.443/0.869 & 0.434/0.863 & 0.504/0.886\\
& & 10-NPO  & \textbf{0.630}/0.885 & \underline{0.627}/\textbf{0.888} & 0.624/0.862 & 0.473/0.851 & 0.480/0.875 & 0.616/\underline{0.887}\\
& & 10-RT   & \underline{0.602}/0.915 & 0.597/\textbf{0.933} & \textbf{0.603}/\underline{0.930} & 0.435/0.878 & 0.435/0.863 & 0.561/0.896\\
\cline{3-9}
& & 20-DPO  & \underline{0.561}/\textbf{0.908} & 0.543/0.899 & \textbf{0.563}/\underline{0.903} & 0.377/0.851 & 0.395/0.867 & 0.531/0.895\\
& & 20-GA   & 0.600/0.892 & 0.595/\textbf{0.914} & \underline{0.601}/\underline{0.905} & 0.436/0.858 & 0.437/0.869 & \textbf{0.605}/0.887\\
& & 20-NPO  & \textbf{0.642}/\underline{0.883} & \underline{0.637}/\textbf{0.886} & 0.635/0.871 & 0.492/0.878 & 0.475/0.861 & 0.632/0.876\\
& & 20-RT   & \textbf{0.597}/\underline{0.909} & 0.588/\textbf{0.923} & \underline{0.592}/0.903 & 0.429/0.886 & 0.427/0.870 & 0.581/0.905\\
\cline{3-9}
& & 30-DPO  & \underline{0.565}/0.897 & 0.545/\textbf{0.918} & \textbf{0.575}/0.892 & 0.390/0.845 & 0.390/0.848 & 0.551/\underline{0.902}\\
& & 30-GA   & \textbf{0.597}/0.890 & \underline{0.595}/\underline{0.896} & \underline{0.595}/\textbf{0.899} & 0.436/0.863 & 0.443/0.888 & 0.594/0.881\\
& & 30-NPO  & \textbf{0.636}/\underline{0.953} & 0.625/\underline{0.953} & \underline{0.629}/\textbf{0.954} & 0.486/0.925 & 0.483/0.939 & \underline{0.629}/0.942\\
& & 30-RT   & \underline{0.591}/0.922 & 0.584/0.947 & \textbf{0.592}/\textbf{0.961} & 0.429/\underline{0.953} & 0.426/0.928 & 0.581/0.912\\
\hline
\multirow{9}{*}{TOFU} & \multirow{9}{*}{\makecell[c]{tofu-ft-llama2-7b\\(\textit{forget01}: 0.550/0.923 \\ \textit{forget05}: 0.538/0.911)}} &
forget01-KL  & \underline{0.424}/\textbf{0.792} & 0.329/0.762 & \textbf{0.455}/0.747 & 0.361/0.755 & 0.338/\underline{0.780} & 0.415/0.735\\
& & forget01-GA  & 0.418/0.739 & 0.335/0.744 & \textbf{0.438}/\textbf{0.780} & 0.374/0.752 & 0.326/\underline{0.765} & \underline{0.419}/0.708\\
& & forget01-GD  & \underline{0.436}/0.771 & 0.341/0.763 & \textbf{0.456}/\textbf{0.783} & 0.385/\underline{0.777} & 0.326/0.655 & 0.419/0.699\\
\cline{3-9}
& & forget05-IDK  & \underline{0.195}/\textbf{0.675} & 0.177/0.635 & 0.217/\underline{0.655} & 0.189/0.632 & 0.182/0.581 & \textbf{0.223}/0.535\\
& & forget05-NPO  & 0.251/0.501 & 0.248/0.488 & 0.254/\textbf{0.534} & \textbf{0.258}/0.514 & 0.252/\underline{0.516} & \underline{0.256}/0.398\\
& & forget10-NPO  & \underline{0.171}/0.358 & 0.166/0.358 & 0.159/0.372 & 0.162/\textbf{0.397} & 0.162/\underline{0.376} & \textbf{0.173}/0.311\\
& & forget10-AltPO & \underline{0.287}/0.578 & 0.275/\underline{0.583} & \textbf{0.297}/\textbf{0.623} & 0.274/0.570 & 0.267/0.564 & 0.284/0.487\\
& & forget05-SimNPO & 0.246/0.447 & 0.232/0.445 & \underline{0.264}/\textbf{0.482} & 0.243/0.463 & 0.235/\underline{0.469} & \textbf{0.266}/0.428\\
& & forget10-SimNPO & 0.177/0.383 & 0.177/0.368 & \textbf{0.209}/0.431 & 0.188/\textbf{0.473} & 0.179/\underline{0.438} & \underline{0.205}/0.346\\
\hline
WHP & \makecell[c]{Llama-2-7b-chat-hf\\(0.865/1.000)} & - & 0.434/0.944 & 0.487/\underline{0.963} & \textbf{0.545}/\textbf{0.997} & 0.493/0.946 & 0.485/0.879 & \underline{0.505}/0.913 \\
\bottomrule
\end{tabular}
\caption{Evaluation of model performance using Top-K sampling. We report both the average and maximum ROUGE scores of the sampled outputs, formatted as \textit{Average/Maximum}. The model performance prior to unlearning is shown in parentheses beneath the base model name. We mark the best and second best performance with \textbf{bold} and \underline{underline}, respectively.}
\vspace{-10pt}
\label{tab:model_perf_sampling}
\end{table*}

\textbf{Preference based unlearning methods are more effecitve:} Our analysis of model outputs reveals that preference-based algorithms are generally more effective. Unlike the IDK approach, which responds with a refusal or uncertainty, preference-based algorithms substitute the original ground-truth knowledge with plausible alternative answers. This strategy enhances the model's robustness against knowledge extraction through auditing, as it avoids directly signaling the absence of information and instead provides a coherent and altered response.

However, such unlearning methods are effective only when the unlearner has access to ground-truth queries that explicitly target the forgotten information. In other words, the model performs well when asked direct questions like \textit{“Who is A?” → “B”}, but struggles with inverse or paraphrased formulations such as \textit{“Who is B?” → “He is A.”}. We provide illustrative examples in Appendix.

\textbf{Generation with Sampling:} Given the auto-regressive generative nature and inherent randomness of LLM outputs, we further evaluate the effectiveness of unlearning algorithms through generation-based sampling. Specifically, we set the temperature to 2 and apply Top-K sampling with $K=40$ to promote diverse outputs for the baseline methods. For each query, we sample 50 responses with a maximum of 64 new tokens. We report both the average and the maximum ROUGE scores across all sampled responses in Table~\ref{tab:model_perf_sampling}.

For the RWKU benchmark, we observe minimal variation in average ROUGE scores across most methods, with the exception of GA. This aligns with our earlier findings regarding the degradation in model utility introduced by gradient ascent-based unlearning. However, the maximum ROUGE score among the sampled responses often exceeds 0.80, suggesting that knowledge acquired during pretraining remains difficult to fully remove, especially when the original pretraining data is inaccessible. Similar patterns are observed in the WHP benchmark.

In contrast, for the TOFU dataset, the maximum ROUGE score after unlearning reaches only around 0.60. We attribute this to the availability of the fine-tuning synthetic dataset during unlearning, which includes all information related to the target fictitious authors. This direct access to ground-truth knowledge allows the unlearning algorithm to more effectively erase relevant information, leading to more complete unlearning outcomes.

\section{Conclusion}
In this work, we proposed an auditing framework for machine unlearning in LLMs, where we evaluate the existing unlearning algorithm. Besides, we propose an auditing algorithm based on activation perturbation to extract model knowledge. We observe that the existing preference-based unlearning methods are more robust against knowledge extraction methods than refusal-based methods. Also, more research should be conducted regarding the challenge of removing knowledge gained during the pretraining stage.

\bibliographystyle{plainnat}
\bibliography{neurips_2025}


\newpage
\appendix

\appendix
\begin{algorithm*}[t]
\caption{Activation Perturbation-based Auditing (ActPert)}
\small
\label{algo:actpert}

\textbf{Perturbation Computation}
\begin{algorithmic}[1]
\STATE \textbf{Input:} Unlearned model $f_u$, query $q$, number of perturbations $n_p$, noise scale $\gamma$
\STATE Tokenize $q$ and compute the embeddings: $E_q \gets f_u.\text{embedding}(T(q))$.
\STATE Identify the token indices $I_u$ related to unlearning target .
\FOR{$n = 1$ to $n_p$}
    \STATE Initialize $\hat{E}_q^{(n)} \gets \text{Clone}(E_q)$
    \FOR{$i \in I_u$}
        \STATE Sample noise: $\Delta_{d_m} \sim \mathcal{N}(0, I_{d_m})$ \hfill \COMMENT{Embedding dimension: $d_m$}
        \STATE Perturb embedding: $\hat{E}_q^{(n)}[i] \gets \hat{E}_q^{(n)}[i] + \gamma \cdot \Delta_{d_m}$
    \ENDFOR
    \STATE Feed $\hat{E}_q^{(n)}$ into $f_u$ and record $l$-th layer outputs: $\hat{A}_l^{(n)}$
\ENDFOR
\STATE Feed original $E_q$ into $f_u$, record $l$-th layer outputs: $A_l$
\FOR{$l = 1$ to $L$}
    \STATE Compute perturbation: $\delta_l \gets A_l - \frac{1}{n_p} \sum_{n=1}^{n_p} \hat{A}_l^{(n)}$
\ENDFOR
\end{algorithmic}

\vspace{0.5em}
\textbf{Inference with ActPert}
\begin{algorithmic}[1]
\STATE \textbf{Input:} Unlearned model $f_u$, query $q$, activation perturbations $\{\delta_l\}_{l=1}^L$
\WHILE{generated token $t$ is not [EOS]}
    \STATE Feed $q$ into $f_u$, injecting $\delta_l$ into layer activations at each $l$
    \IF{Greedy decoding}
        \STATE $t \gets \arg\max(f_u(q))$
    \ELSE
        \STATE $t \gets \text{sample from top-k}(f_u(q))$
    \ENDIF
    \STATE Append $t$ to query: $q \gets q + t$
\ENDWHILE
\STATE \textbf{Return} $q$
\end{algorithmic}
\end{algorithm*}

\section{Related Work}
\subsection{Machine Unlearning in LLMs}
Machine unlearning has garnered significant attention in the context of LLMs. Various approaches for targeted knowledge removal have been proposed: ~\citet{eldan2023harrypotter} removed Harry Potter-related knowledge by finetuning LLMs on corpora with replaced keywords. ~\citet{zhang2024negative} proposed to steer model preferences in the negative direction to reduce memorization. ~\citet{wang2024rkld} used reversed knowledge distillation to eliminate personal information. ~\citet{feng2024fine} introduced a reweighted gradient ascent method for unlearning, and ~\citet{pawelczyk2023context} utilized in-context unlearning examples. ~\citet{liu2024large, bhaila2024soft} adapted input embeddings associated with the unlearning target, while~\citet{li2024wmdp, tamirisa2024toward, huu2024effects, ashuach2024revs} proposed interventions in the model activation space. From a safety perspective, several studies have targeted the unlearning of harmful or undesirable model behaviors~\cite{zhang2024safe, yao2023large, liu2024towards}, while others have addressed backdoor unlearning~\cite{liang2024unlearning, liu2024efficient} to enhance model integrity.

\subsection{Evaluation of LLM Unlearning}
While advances have been made in developing machine unlearning algorithms for LLMs, rigorous methodologies for auditing the efficacy of unlearning remain understudied. Recent work has demonstrated that unlearned knowledge can be recovered using techniques such as adversarial soft token attacks~\cite{schwinn2024soft} or few-shot in-context prompting~\cite{doshi2024doesunlearningtrulyunlearn}. When model parameters are accessible, recovery can also be achieved through model quantization~\cite{zhang2024does} or retraining on partially unlearned datasets~\cite{lucki2024adversarial, hu2024jogging}. ~\citet{lynch2024eight} compared eight unlearning evaluation methods and found persistent similarity in the latent representations of unlearned models. ~\citet{shi2024muse} conducted a comprehensive analysis across six evaluation dimensions using news and book datasets, highlighting the fragility of unlearning to paraphrasing and unrelated data. ~\citet{qi2024unrelateddata} showed that fine-tuning on unrelated data could inadvertently restore previously removed knowledge. Several benchmarks have been developed to evaluate the existing unlearning algorithms. ~\citet{maini2024tofu} proposed a benchmark centered on fictitious author profiles, while \citet{jin2024rwku} introduced the Real-World Knowledge Unlearning (RWKU) benchmark, which targets knowledge removal of 200 well-known individuals. Complementary to this, the WDMP benchmark~\cite{li2024wmdp} focuses on the unlearning of sensitive knowledge related to biosecurity and cybersecurity threats.

\section{Example Model Outputs}

\begin{tcolorbox}[colback=gray!5!white,colframe=gray!75!black]
\textbf{Examples of Gradient Ascent}:

\textcolor{blue}{\textbf{Query}}: Basil Mahfouz Al-Kuwaiti's father was a \_\_\_.

\textcolor{blue}{\textbf{GT}}: Florist

\textcolor{blue}{\textbf{$f_u$} (GA/forget-1\%)}: Barber

\textcolor{blue}{\textbf{$f_u$} (GA/forget-5\%)}: TTTTTTTTTTT
\end{tcolorbox}

\begin{tcolorbox}[colback=gray!5!white,colframe=gray!75!black]
\textbf{Examples of Preference-based Unlearning}:

\textcolor{blue}{\textbf{Query}}: Hina Ameen primarily contributes to the \_\_\_ genre. 

\textcolor{blue}{\textbf{GT}}: Geology.

\textcolor{blue}{$f_u$ (IDK)}: I'm not programmed to know that.

\textcolor{blue}{$f_u$ (NPO)}: Genre of literary fiction.
\end{tcolorbox}

\begin{tcolorbox}
\textbf{Examples of Order Changing}:

\textcolor{blue}{\textbf{Query}}: Who is the headmaster of Hogwarts in Harry Potter series?

\textcolor{blue}{$f_u$}: There is no character named "Headmaster of Hogwarts" in the Harry Potter series.

\textcolor{blue}{\textbf{Query}}: Who is Albus Dumbledore in Harry Potter series?

\textcolor{blue}{$f_u$}: Albus Dumbledore is the headmaster of Hogwarts...
\end{tcolorbox}

\begin{table*}[t]
\centering
\small
\begin{tabular}{ll}
\toprule
\textbf{Symbol} & \textbf{Meaning} \\
\midrule
$ \mathcal{L}_{\text{DPO}} $ & DPO loss function \\
$ \mathcal{D}_u $ & Unlearning dataset \\
$ x $ & Input sample \\
$ y_w, y_l $ & Preferred and less preferred responses \\
$ f_u(y \mid x) $ & Output probability from unlearned model \\
$ f_{ft}(y \mid x) $ & Output probability from fine-tuned model \\
$ \beta $ & Temperature scaling factor \\
$ \sigma(\cdot) $ & Sigmoid function \\
\bottomrule
\end{tabular}
\caption{Symbol definitions for the DPO loss function.}
\label{tab:dpo_symbol_table}
\end{table*}

\begin{table*}[t]
\centering
\small
\begin{tabular}{ll}
\toprule
\textbf{Symbol} & \textbf{Meaning} \\
\midrule
$ x $ & Input query \\
$ y_l $ & Ground-truth label (target to forget) \\
$ \mathcal{D}_u $ & Unlearning dataset \\
$ f_u(y_l \mid x) $ & Output probability from the unlearned model \\
$ f_{ft}(y_l \mid x) $ & Output probability from the original (fine-tuned) model \\
$ \beta $ & Scaling factor for preference shift \\
$ \sigma(\cdot) $ & Sigmoid function: $\sigma(z) = \frac{1}{1 + e^{-z}}$ \\
$ \log \sigma(\cdot) $ & Log-likelihood used as loss for optimization \\
\bottomrule
\end{tabular}
\caption{Explanation of symbols used in the NPO loss function.}
\label{tab:npo_symbol_table}
\end{table*}

\begin{table*}[t]
\centering
\small
\begin{tabular}{ll}
\toprule
\textbf{Symbol} & \textbf{Meaning} \\
\midrule
$ \theta_u $ & Model \\
$ A_k $ & Activation at layer $k$ for the original input \\
$ \hat{A}_k^i $ & Activation at layer $k$ for the $i$-th perturbed input \\
$ \delta_k $ & Difference activation between original and perturbed in layer $k$ \\
$ n_p $ & Number of noise samples \\
$ d_m $ & Dimension of one embedding \\
$ X \sim \mathcal{N}(0, 1) $ & Gaussian distribution \\
\bottomrule
\end{tabular}
\caption{Symbol definitions for model and perturbation-related variables.}
\label{tab:model_symbol_table}
\end{table*}

\section{Prompts for Dataset Generation}
\label{sec:appendix}
In this section, we provide the prompts used to generate the datasets for auditing unlearned in \textit{WHP} and \textit{TOFU} benchmarks.

Prompt for \textit{WHP}:
\begin{tcolorbox}
Please generate 35 short, fact-based question-and-answer pairs related to the Harry Potter series. Each question should be clearly answerable with a brief response (e.g., a name, place, object, or short phrase). Ensure that all questions are specific to the Harry Potter universe. Provide both the question and its corresponding answer for each pair.
\end{tcolorbox}

Prompt for \textit{TOFU}:
\begin{tcolorbox}
Please rewrite the following question-and-answer pair into fill-in-the-blank format. Each blank should be clearly answerable with a brief response (e.g., a name, place, object, or short phrase). 
\end{tcolorbox}

\section{Implementation Details for ActPert}
In this section, we provide further details about the hyperparameters for the proposed method. Specifically, we set the layer index for computing the activation difference as 12 and set the noise intensity as 0.01. We observe that using shallow layers or larger noise intensity would significantly reduce the model utility and make model outputs random characters, while using deeper layers would degrade the auditing performance. 

\begin{table}[ht]
\centering
\small
\setlength{\tabcolsep}{4.25pt}
\begin{tabular}{c|c|c|c|c|c|c|c}
\toprule
Algo. & Base & 6 & 9 & 12 & 15 & 18 & 21  \\
\hline
forget01-KL  & 0.503 & 0.426 & 0.461 & 0.526 & 0.503 & 0.510 & 0.491 \\
forget01-GA  & 0.503 & 0.394 & 0.435 & 0.590 & 0.572 & 0.518 & 0.543 \\
forget01-GD  & 0.525 & 0.412 & 0.446 & 0.568 & 0.541 & 0.509 & 0.531 \\
forget05-IDK  & 0.212 & 0.184 & 0.197 & 0.253 & 0.237 & 0.268 & 0.226 \\
\bottomrule
\end{tabular}
\caption{Evaluation of audited model performance using ActPert across different layer indices.}
\label{tab:ablation_layer}
\end{table}

\begin{table}[htb]
\centering
\small
\setlength{\tabcolsep}{4.25pt}
\begin{tabular}{c|c|c|c|c|c|c}
\toprule
Algo. & Base & 0.002 & 0.005 & 0.01 & 0.02 & 0.04 \\
\hline
forget01-KL  & 0.503 & 0.517 & 0.521 & 0.526 & 0.401 & 0.298 \\
forget01-GA  & 0.503 & 0.562 & 0.597 & 0.590 & 0.435 & 0.302\\
forget01-GD  & 0.525 & 0.532 & 0.551 & 0.568 & 0.426 & 0.259 \\
forget05-IDK  & 0.212 & 0.256 & 0.276 & 0.253 & 0.204 & 0.128 \\
\bottomrule
\end{tabular}
\caption{Evaluation of audited model performance using ActPert with different noise intensity.}
\label{tab:ablation_noise}
\end{table}


\newpage

\end{document}